\documentclass[sigplan,screen,nonacm]{acmart}
\begin{document}

\title{DYPLODOC: Dynamic Plots for Document Classification}

\author{Anastasia Malysheva}
\affiliation{%
  \institution{Open Data Science}
  \city{Moscow}
  \country{Russia}
}
\email{amalysheva42@yandex.com}

\author{Alexey Tikhonov}
\affiliation{%
  \institution{Yandex}
  \city{Berlin}
  \country{Germany}}
\email{altsoph@gmail.com}

\author{Ivan P. Yamshchikov}
\affiliation{%
  \institution{LEYA Lab, Yandex, Higher School of Economics}
  \city{St.Petersburg}
  \country{Russia}
}
\email{ivan@yamshchikov.info}

\renewcommand{\shortauthors}{Malysheva, et al.}

\begin{abstract}
  Narrative generation and analysis are still on the fringe of modern natural language processing yet are crucial in a variety of applications. This paper proposes a feature extraction method for plot dynamics. We present a dataset that consists of the plot descriptions for thirteen thousand TV shows alongside meta-information on their genres and dynamic plots extracted from them. We validate the proposed tool for plot dynamics extraction and discuss possible applications of this method to the tasks of narrative analysis and generation.
\end{abstract}



\maketitle

\section{Introduction}

Natural Language Generation (NLG) is one of the areas within Natural Language Processing (NLP). Deep learning enabled various generative applications where generated texts are short and constrained by the context. These could be dialogue responses \cite{li2016deep,li2019dialogue}, traditional or stylized poetry \cite{zhang2014chinese,wang2016chinese,tikhonov2018guess}, end-to-end author or source stylization \cite{hylsx,tikhonov2019style,syed2020adapting}. One of the bottle-necks that holds modern Natural Language Processing (NLP) from the generation of longer texts is the concept of {\em narrative}.  There are constant attempts to generate longer blocks of text, such as \cite{kedziorski2019understanding} or \cite{agafonova2020paranoid}, yet they succeed under certain stylistic and topical constraints that exclude the problem of narrative generation altogether \cite{van2019narrative}.

Though philosophers and linguists try to conceptualize the notions of plot, narrative arc, action, and actor for almost a century \cite{shklovsky1925theory,propp1968morphology,van1976philosophy}, few of these theoretical concepts could be instrumental for modern NLP. \citet{ostermann2019mcscript2} present a machine comprehension corpus for the end-to-end evaluation of script knowledge. The authors demonstrate that though the task is not challenging to humans, existing machine comprehension models fail to perform well, even if they make use of a commonsense knowledge base. Despite these discouraging results, there are various attempts to advance narrative generation within the NLP community, see \cite{fan2019strategies,ammanabrolu2020story}. For a detailed review of approaches to narrative generation, we address the reader to \cite{kybartas2016survey}, here we just mention several ideas that are relevant within the scope of this paper.

One line of thought is centered around some form of a multi-agent system with various constraints that keep generated texts within some proximity of human-written narrative. For example, \citet{theune2004emotional} present a multi-agent framework for automatic story generation where plots are automatically created by semi-autonomous characters while the resulting plot is converted into text by a narrator agent. \citet{walker2016m2d} experiments with a dialogic form of a story: a deep representation of a story is converted into a dialog that unfolds between two agents with distinctive stylistic characteristics. Such multiagent-based systems show a lot of potential yet do not leverage the existing datasets that include an extensive amount of narratives written by humans.

Another line of research tries to use some NLP methods and statistical procedures to systematize existing narratives using some form of synthetic features that could in some form describe the narrative structure of a given story. \citet{y2007employing} represents a story as a cluster of emotional links and tensions between characters that progress over story-time. \citet{schmidt2015plot} introduces a concept of narrative arc in a context of large-scale text mining. The author conceptualizes plot structure as a path through a multidimensional space derived from a topic model. This approach allows us to see plot dynamics within any given book or show in detail, yet it is harder to generalize. Topic modeling tends to provide profoundly different sets of topics across genres. When aggregating across all topics within a dataset one gets a high dimensional space that is harder to interpret. For example, similar narrative arcs of a detective story would look different in terms of topic modeling if the same plot takes place in Victorian England or on Mars. One would like to keep the concept of a narrative arc yet to make it more abstract in order to be able to carry comparative narrative research across genres and settings. \citet{mathewson2019shaping} develop ideas that address this challenge to a certain extent. The authors mathematize a concept of narrative arc that depicts the evolution of an agent's belief over a set of the so-called {\em universes}. The authors estimate these probabilities with pre-trained probabilistic topic classifiers and combine these estimates with a multi-agent dialogic approach. They demonstrate that the agent that tracks the dynamics of such narrative arcs generates more engaging dialog responses.

This paper shows that the concept of the narrative arc could be applied to a vast variety of narratives in natural languages and does not have to be limited to the setting of a dialogue system. The contribution of this paper is two-fold: (1) we provide a large dataset of longer storylines that could be used for narrative research and (2) use it to demonstrate how the notion of the narrative arc could be applied to longer texts and provide meaningful insights into the structure of the story. This structural insight could be useful both for quantitative research of narrative dynamics and as synthetic features for the generation of narrative structures in the future.

\section{Data}

TVmaze\footnote{https://www.tvmaze.com/} is a huge database with the data on different TV shows that include the storyline as well as certain meta information. Using TVmaze API we have obtained 300+ Mb of data that could be found online as a part of the DYPLODOC package\footnote{https://github.com/AnastasiaMalysheva/dyplodoc}. The dataset includes 13 813 TV shows and 21 962 seasons that consist of more than 300 000 episodes. Every episode in the dataset is annotated with a condensed description of the main storyline and the narrative structure of a season is defined as a sequence of such descriptions. The data includes main characters, that could be extracted with the NER algorithm, the majority of texts are centered around most reportable events \cite{ouyang2015modeling}, and the narrative is split into clear steps, certain 'quanta' of a story. Almost every season in TVmaze has a tag of a genre, such as (\verb"Horror", Food, Drama, etc). We hope that the publication of the dataset could facilitate further quantitative research of narrative structures.



\section{Plot Dynamics Extraction Tool}

Here we leverage the fact that episode descriptions in the TVmaze dataset form seasons and demonstrate how one could build the narrative arc of a season by constructing artificial features that would to a certain degree describe the season's plot dynamics. We call this method Plot Dynamics Extraction (PDE) Tool.

Let us regard a specific TV show $X$ as a set of $N$ seasons $X: \{X_1, X_2, ..., X_N\}$. Every season $X_i$ could be represented as a sequence of episodes $X_i: \{y_1, y_2, ..., y_M \}$. Every $y_i$ is a description of one particular episode. 

We investigated the plot dynamics using two different pre-trained probabilistic topic classifiers. However, unlike \cite{schmidt2015plot} we choose fixed topics that form a universal state space of narrative arcs. One could discuss if the provided list of dimensions is complete, yet we believe that two classifiers are sufficient to extract non-trivial narrative arcs out of a variety of texts: 
\begin{itemize}
    \item Classifier of movie genres trained on Wikipedia film synopses provided in \cite{hoang2018predicting}. The labels of genres include: \verb"Comedy",  \verb" Drama",  \verb" Western",  \verb" Adventure", \verb" Animation", \verb" Action", \verb" Thriller", \verb" Family", \verb" Romance", \verb" Fantasy", \verb" Horror", \verb" History", \verb" Music", \verb" Sci-Fi", \verb" War", \verb" Crime", \verb" Musical", \verb" Biography", \verb" Mystery", \verb" Sport".
    \item DeepMoji: Deep neural network that was trained on a corpus of text that included 64 emojis, see \cite{felbo2017using}. In their blogpost\footnote{https://medium.com/@bjarkefelbo/what-can-we-learn-from-emojis-6beb165a5ea0} the authors suggest hierarchic aggregation of 64 emojis. Using this aggregation we limit ourselves to 11 labels in this work. These labels include: \verb"Love", \verb"Happy", \verb"Wink", \verb"Deal", \verb"Force", \verb"Eyes", \verb"Fear", \verb"Mad", \verb"Sad", \verb"Music", \verb"Misc".
\end{itemize}
Thus every narrative arc is a trajectory on a union of two simplexes: a 20-simplex of genres and an 11-simplex of emotions. For the vast majority of the cases, the classifier mostly detects one or two most dominant genres and topics respectively, limiting the meaningful development of a given narrative art to one plane on each simplex. 

Similarly to \cite{mathewson2019shaping} we propose to regard a finite set of universes, $\mathcal{U}$. One could develop a universe model to assess the compatibility of an episode $y_i$ with a given universe, $u \in \mathcal{U}$. Given such a model, one could update the posterior universe distribution over a sequence of episodes $y_i$. For each universe $u$, the universe model assigns a likelihood $p(y_i | y_{1:i-1}, u)$ to an episode $y_i$, conditioned on previous episodes $y_{1:i-1}$ of the season. Introducing $z(u)$ as the prior distribution over universes, the conditional probability $z(u|y_i)$ could be defined as follows $z(u|y_i) = z(u) \times \frac{z(y_i|u)}{z(y_i)}$.

Using this notation \cite{mathewson2019shaping} obtain the following approximation

\begin{equation} \label{eq:u}
    p(u|y_{1:i}) = p(u|y_{1:i-1}) \times \frac{z(u|y_i)}{\sum_{u'} p_{i-1}(u'|y_{1:i-1})z(u'|y_i)}
\end{equation}

In order to guarantee that the sequence of $p(u|y_{1:i})$ forms a series of coordinates on the union of 20-simplex of genres and 11-simplex of emoji, we introduce two additional smoothing procedures. First, we control $z(u)$ distribution update so that it does not degenerate into zero, since in this case Equation (\ref{eq:u}) becomes obsolete. We also use exponential smoothing average instead of standard multiplicative update. 

\begin{figure}[t]
\centering
     \includegraphics[scale=0.45]{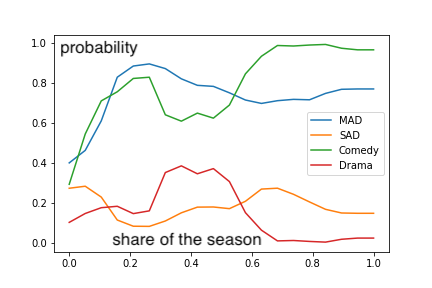}
  \caption{Dynamics of the fourth season of 'BoJack Horseman'. Tags from the simplex of emotions are denoted with capital letters.}
   \label{fig:bojack}
\end{figure}

Figure \ref{fig:bojack} illustrates the resulting dynamics of one season of 'BoJack Horseman', extracted with a Deepmoji classifier and with a classifier pre-trained on the genres\footnote{PDE Tool is available as a part of DYPLODOC package https://github.com/AnastasiaMalysheva/dyplodoc}. There are four major non-zero components. On the emoji-simplex the narrative is balancing between \verb"MAD" and \verb"SAD", though the former is prevailing throughout the season, the latter peaks in the episode when BoJack's daughter overdoses and is found in the hospital. On the genre-simplex, the narrative is balancing between \verb"Comedy" and \verb"Drama". BoJack Horsemen is a dark comedy animation series so it stands to reason that \verb"Comedy" dominates the season while \verb"Drama" peaks in the second quarter of the season when BoJack spends time with his elderly mother in the nursing home.

\section{Experimenting with PDE Tool}

Though scores of pre-trained deepmoji and genre classifiers have very intuitive interpretation it is hard to claim that they could be useful for any form of comparative narrative research. The dynamic aspect of these constructed features is clear yet it is not obvious that their dynamics are meaningful in any non-trivial sense. We compare dynamic plots extracted from the season description with static meta-information that is included in the TVmaze data\footnote{7911 seasons out of 21962 have no tags so we exclude them from this experiment.}. If different episodes of the season are labeled with a different tag we choose the most frequent tag as the tag for the whole season. If the show was tagged with two values\footnote{We filtered the shows that had more than 2 tags for their genre.}, we treated it like one static tag. For example, 'BoJack Horseman' has \verb"Drama | Comedy" as a genre, thus \verb"Drama", \verb"Comedy" and  \verb"Drama | Comedy" for three separate static tags. 

PDE-tool represents every plot as an array of tags (such as genres or emoticons), in which every tag is represented as a sequence of probability estimations ranging from 0 to 1. We concatenate these representations with extra synthetic features, namely: mean, median, minimum, maximum, and variance for the probability of every tag; area under the curve; the moment of intersection with another tag probability estimation, etc. Thus we represent every text as a multidimensional vector. Using these dynamic based features we clustered the seasons, applying standard K-means clustering over a PCA\footnote{Since the length of the seasons might differ and we are interested in comparative research of the obtained arcs we scale every series to the length of ten episodes either uniformly stretching or compressing the resulting timeline.}. We have carried our experiments across all possible combinations of features, a different number of principal components (50, 100, 200, 300), and a various number of clusters (50, 75, 100, 150, 200, 250, 300). Then we calculated mutual information between obtained clusters and static tags for the corresponding seasons. The higher values of mutual information were obtained for the following setup: synthetic features included statistics and area under the curve for every component; the number of principal components equals 200; the number of clusters for K-means equals 200.

With this setup, the mutual information between clusters based on both classifier combined and static tags was $0.57$, as shown in Table \ref{tab:mi}.  One could see that both the dynamics based on the genre classifier and the emoji classifier-based dynamics allow inferring static genre tags. The combination of two classifiers yields higher mutual information. Dynamic plots extracted by different classifiers differ significantly and go beyond static genre tags.

\begin{table}[ht!]
\centering
\begin{tabular}[t]{lrr}
\hline
                                                                      & \multicolumn{1}{l}{\begin{tabular}[c]{@{}l@{}}MI\end{tabular}} & \multicolumn{1}{l}{Silhouette} \\ \hline
\begin{tabular}[c]{@{}l@{}}Dynamic\\ Genres\end{tabular}              & 0.47                                                                             & 0.23                              \\
\hline
\begin{tabular}[c]{@{}l@{}}Dynamic\\ Emojis\end{tabular}              & 0.36                                                                             & 0.16                              \\
\hline
\begin{tabular}[c]{@{}l@{}}Combined\\ Dynamic\\ Features\end{tabular} & 0.57                                                                   & 0.20\\
\hline
\end{tabular}
\caption{Mutual Information (denoted as MI) between k-means clusters based on the features obtained with plot dynamics extraction and the clusters defined by static genre meta-information alongside with the silhouette for the obtained clusterizations.} \label{tab:mi}
\end{table}
\begin{table}[ht!]
\centering
\begin{tabular}[t]{llr}
\hline
\multicolumn{1}{l}{\begin{tabular}[l]{@{}l@{}}\# of \\ shows \end{tabular}} & \multicolumn{1}{l}{\begin{tabular}[l]{@{}l@{}}Dominant \\ Static Tag\end{tabular}} & \multicolumn{1}{l}{\begin{tabular}[l]{@{}l@{}}\% of \\ Dominant \\ Static Tag\end{tabular}} \\ \hline
274                      &\verb"Comedy | Drama"                                                               & 74                                                                           \\
260                      &\verb"Drama | Fantasy"                                                            & 79                                                                           \\
253                      &\verb"Drama | Comedy"                                                               & 72 \\
\hline
\end{tabular}
\caption{Three biggest clusters in the resulting clusterization that have the highest percentage of a dominant tag.}\label{tab:ten}
\end{table}

\begin{figure*}[t]
     \includegraphics[scale=0.25]{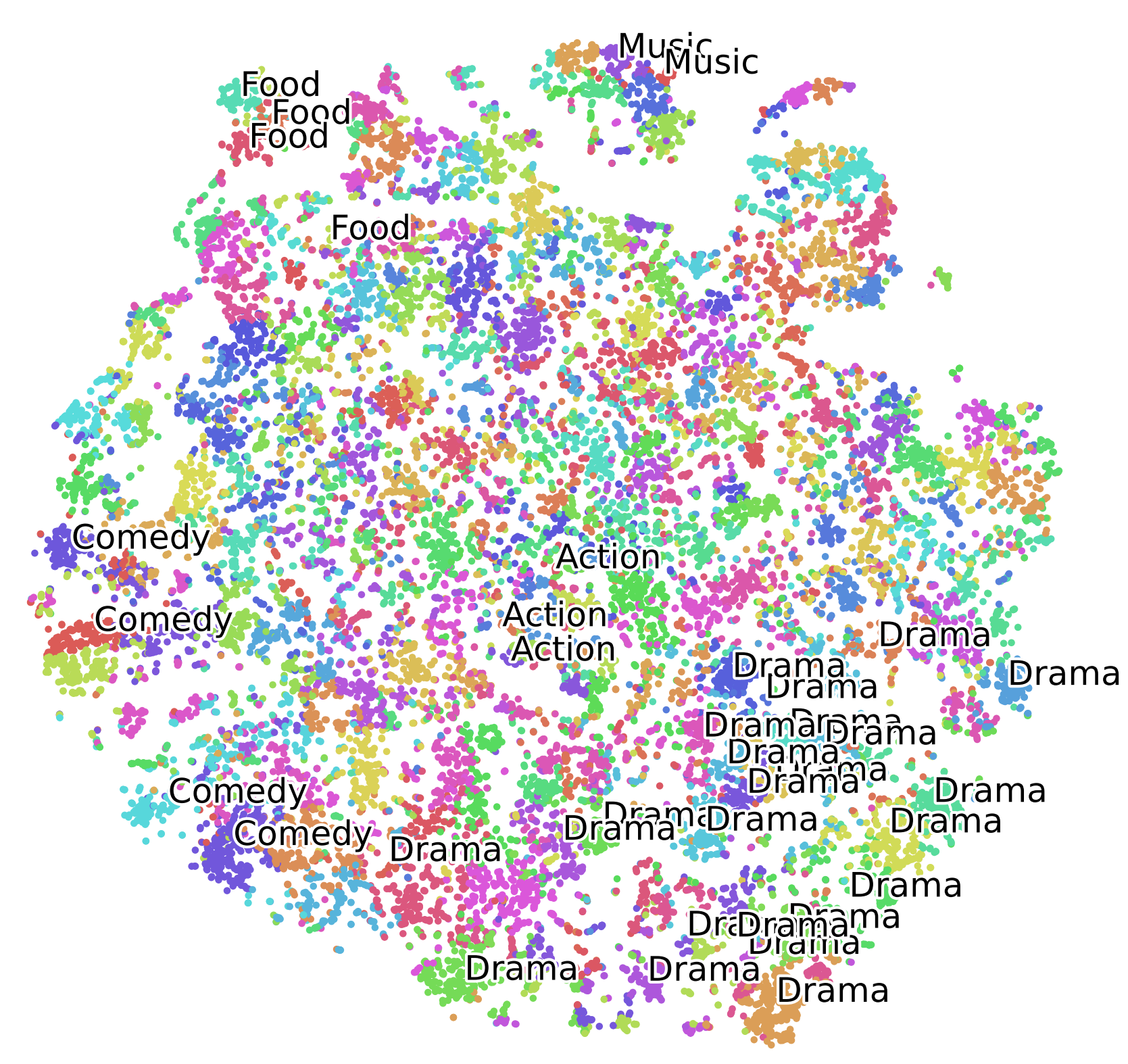}
  \caption{t-SNE visualization of the clustering based on plot dynamics. The labels mark clusters in which the dominant genre tag corresponds to $70 \%$ or more shows in the cluster. }
  \label{fig:clus}
\end{figure*}

\section{Discussion}

Figure \ref{fig:clus} shows the resulting t-SNE (t-distributed stochastic neighbor embedding) visualization, see \cite{van2008visualizing} for details, of the clustering based on plot dynamics. The centers of the clusters are labeled with static genre tags. The average size of a cluster in the resulting clustering is 109 and the average share of the dominant tag is $65 \%$. The smallest of the clusters consists of 9 show seasons, the biggest includes 288 seasons. In some clusters the share of the dominant tag reaches $100 \%$, the lowest possible percentage of the dominant tag is $42 \%$.

Table \ref{tab:clus} describes basic parameters of the cluster distribution. 

\begin{table}[t]
\centering
\begin{tabular}{lrr}
\hline
       & Size & \begin{tabular}[c]{@{}l@{}}Dominant \\ tag \%\end{tabular} \\ \hline
Minimum & 9    & 42                                                         \\
Median  & 107  & 65                                                         \\
Average & 109  & 65                                                         \\
Maximum & 288  & 100                                                       
\end{tabular}
\caption{Minimum, median, average and maximum values for the size and percentage of the dominant tags in the resulting clusterization based on the dynamic plots.}
\label{tab:clus}
\end{table}

\begin{figure}[t]\centering
     \includegraphics[scale=0.45]{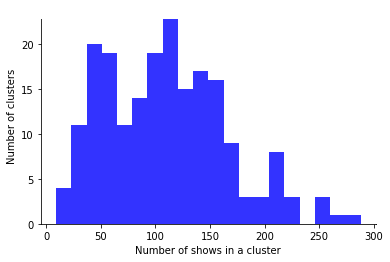}
 \caption{Distribution of the resulting clusters in terms of their sizes.}
\label{fig:clusdis}
\end{figure}

Using synthetic features based on the PDE tool one could partially infer static human labels for season's genre. Table \ref{tab:mi} demonstrates that the dynamic plot structure is non-trivial and could be useful for quantitative research of narrative. We publish extracted plot dynamics along with the TVmaze data to facilitate further research.

Now let us show how the PDE tool could be used for comparative research of narrative. Let us zoom in on the 200 clusters obtained from dynamics plots extracted from the descriptions of the seasons. Table \ref{tab:ten} shows the three biggest resulting clusters with the highest percentage of one static genre tags. 

\begin{figure}[ht]
\centering
     \includegraphics[scale=0.45]{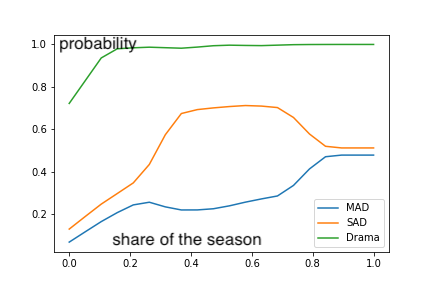}
  \caption{Dynamics of the second season of 'Victoria'. Tags from the simplex of emotions are denoted with capital letters.}
  \label{fig:vic}
\end{figure}

\begin{figure}
\centering
     \includegraphics[scale=0.45]{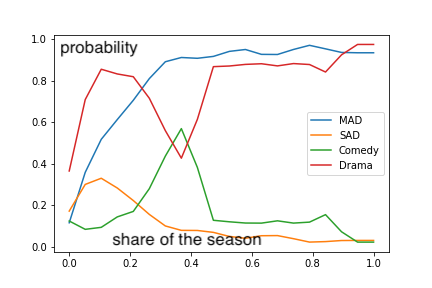}
  \caption{Dynamics of the third season of 'Orange is a new black'. Tags from the simplex of emotions are denoted with capital letters.}
  \label{fig:or}
\end{figure}

The first cluster could be attributed to darker comedies. Figure  \ref{fig:bojack} shown above is a good representative of this cluster. The second cluster could be attributed to historic dramas. The cluster is characterized by the dominance of \verb"Drama" and interplay of \verb"Sad" and \verb"Mad" emojis. Figure \ref{fig:vic} illustrates it with the dynamics of the second season of Victoria. The third cluster is also dominated by \verb"Drama" but with the certain presence of \verb"Comedy" as well as with domination of \verb"Mad" over \verb"Sad", see Figure \ref{fig:or} for an example. These are only a few examples that illustrate how rich is the information provided by the PDE tool and its potential for quantitative analysis of narrative.

\section{Conclusion}

This work makes the following contributions: (1) a universal feature extraction method for plot dynamics; (2) a dataset of natural language plot descriptions; (3) finally, the paper shows that plot dynamics features extracted with the proposed method are meaningful and can shed light on the general structure of the narrative.

We hope that this paper could help to revive the attention of the community to the problems of narrative analysis and generation at the very least by providing data and tools for further exploration. We also hope to use narrative arch extracted with PDE tool as a form of semi-supervision for narrative-oriented generative models. The extracted dynamic plots are presented alongside the textual data. We hope that the proposed tool would inspire further research on plot dynamics and inform the discussions on narrative structure, particularly in the context of narrative generation.

\bibliographystyle{acl_natbib}
\bibliography{bibl}
\end{document}